\documentclass[conference]{IEEEtran}
\IEEEoverridecommandlockouts
\usepackage{cite}
\usepackage{amsmath,amssymb,amsfonts}
\usepackage{algorithm}
\usepackage{algpseudocode}
\usepackage{graphicx}
\usepackage{subcaption}
\usepackage{bm}
\usepackage{tikz}
\usepackage{textcomp}
\usepackage{xcolor}
\def\BibTeX{{\rm B\kern-.05em{\sc i\kern-.025em b}\kern-.08em
    T\kern-.1667em\lower.7ex\hbox{E}\kern-.125emX}}

\newcommand{\R}{\mathbb{R}}
\newcommand{\C}{\mathbb{C}}

\newcommand{\br}{\boldsymbol{\rho}}

\newcommand{\FF}{\mathbf{F}}

\newcommand{\GG}{\mathbf{G}}
\newcommand{\RR}{\mathbf{R}}
\newcommand{\HH}{\mathbf{H}}
\newcommand{\dd}[1]{\mathbf{d}_{#1}}
\newcommand{\cc}{\mathbf{c}}
\newcommand{\bb}{\mathbf{b}}
\newcommand{\zz}{\mathbf{z}}
\newcommand{\xx}{\mathbf{x}}
\newcommand{\bx}{\bm{x}}
\newcommand{\uu}{\mathbf{u}}

\newcommand{\hc}{\mathbf{\hat{c}}}

\newcommand{\AAA}{\mathbf{A}}

\begin{document}

    \title{Online Sparse Synthetic Aperture Radar Imaging
    \thanks{This work was supported by the Air Force Office of Scientific Research
(AFOSR) under the agreement FA9550-23-1-0604.}
    }

\makeatletter
\newcommand{\linebreakand}{%
  \end{@IEEEauthorhalign}
  \hfill\mbox{}\par
  \mbox{}\hfill\begin{@IEEEauthorhalign}
}
\makeatother

\author{\IEEEauthorblockN{Conor Flynn}
\IEEEauthorblockA{Department of Electrical, Computer, \\and Systems Engineering\\
\textit{Rensselaer Polytechnic Institute}\\
110 8th Street, Troy, NY, 12180 USA \\
flynnc3@rpi.edu}
\and
\IEEEauthorblockN{Radoslav Ivanov}
\IEEEauthorblockA{Department of Computer Science\\
\textit{Rensselaer Polytechnic Institute}\\
110 8th Street, Troy, NY, 12180 USA \\
ivanor@rpi.edu}
\and
\IEEEauthorblockN{Birsen Yaz\i c\i}
\IEEEauthorblockA{Department of Electrical, Computer, \\and Systems Engineering\\
\textit{Rensselaer Polytechnic Institute}\\
110 8th Street, Troy, NY, 12180 USA \\
yazici@ecse.rpi.edu}
}

\maketitle

\begin{abstract}
With modern defense applications increasingly relying on inexpensive, autonomous drones, lies the major challenge of designing computationally and memory-efficient onboard algorithms to fulfill mission objectives. This challenge is particularly significant in Synthetic Aperture Radar (SAR), where large volumes of data must be collected and processed for downstream tasks. We propose an online reconstruction method, the Online Fast Iterative Shrinkage-Thresholding Algorithm (Online FISTA), which incrementally reconstructs a scene with limited data through sparse coding. Rather than requiring storage of all received signal data, the algorithm recursively updates storage matrices for each iteration, greatly reducing memory demands. Online SAR image reconstruction facilitates more complex downstream tasks, such as Automatic Target Recognition (ATR), in an online manner, resulting in a more versatile and integrated framework compared to existing post-collection reconstruction and ATR approaches.
\end{abstract}

\section{Introduction}
Given the shift of modern defense applications to inexpensive, autonomous drones comes the challenge of designing resources-efficient algorithms that maintain performance comparable to legacy systems \cite{coffey2002emergence, lehtotextordmasculine2021small}. Such a challenge is emphasized in Synthetic Aperture Radar (SAR), where many pulses are transmitted over a scene with the back-scattered signals recorded, often resulting in large volumes of data \cite{curlander1991synthetic}. Post processing techniques, such as noise removal and Automatic Target Recognition (ATR), also commonly are done post data collection as using them during collection requires online reconstruction and would be too computationally inefficient in current systems \cite{singh2016analysis, zeng2013sar, bhanu2007automatic}. To address these challenges we propose an online reconstruction approach that requires fewer pulses and less memory storage. While the proposed algorithm could also be integrated to online parameter tuning and ATR, these applications are beyond the scope of this paper.

Compressive sensing, the underlying methodology behind our proposed algorithm, attempts to classify a scene based on the mapping between a set of sparse coefficients and a predetermined dictionary \cite{baraniuk2007compressive}. This set of predicted coefficients can then be used in reference with the dictionary to reconstruct an image \cite{hillar2015can}. Sparse coding is highly effective in areas such as sparse SAR, where we aim to reconstruct an image using a limited number of pulses by exploiting the underlying sparsity of the signals \cite{xu2022sparse, souganlui2014dictionary}. Fourier-based reconstruction techniques, such as BackProjection (BP) \cite{jakowatz2008beamforming}, are unfit for this problem, as the quality of the reconstructed image is proportional to the size of the data collection manifold. While an improvement in this aspect over Fourier-based algorithms, sparse SAR algorithms require increased computational cost to compensate for the small data collection manifold.

A common optimization algorithm used in sparse coding is the Iterative Shrinkage-Thresholding Algorithm (ISTA) \cite{ista}, though its high computational cost makes it unsuitable to run at every slow-time iteration. Fast ISTA (FISTA) \cite{fista} attempts to resolve the number of iterations by introducing a momentum coefficient, though it still is too strenuous for online sparse SAR reconstruction. Online variants of these algorithms have been proposed for signal processing \cite{mairal2010online, lu2013online, gao2014online, wang2018scalable}, as well as ones specifically for medical image reconstruction \cite{mason2017deep, souganlui2014dictionary, moore2019online, cao2024sodl}. Our proposed algorithm shares similarities between these algorithms, such as in carrying over coefficients between iterations and limiting data storage requirements when processing. However, our algorithm is developed specifically for sparse SAR through inclusion of the SAR forward model and derivation of sufficient statistics as outlined in Section~\ref{sec:sparse-sar}.

Dictionary formation is another key aspect of compressive sensing, where we determine the optimal atoms to comprise the dictionary of which yields the most accurate reconstructions. Most of the aforementioned algorithms learn the dictionaries, often built upon Learned ISTA (LISTA) \cite{lista} and the K-SVD Algorithm \cite{aharon2006k}, which learn the atoms of the dictionary as data is collected. While useful in most cases where there is sufficient prior training data, SAR often operates in unfamiliar, complex scenes where unknown targets may not be well-represented in training data \cite{flynn2025comparability}. Therefore, we use a over-complete, static (non-learned) dictionary comprised of edgelets, allowing for a more definitive sparse representation of targets which is especially useful in ATR. While beneficial in that no training is required before running the system on an unknown scene, it does come with the condition that the predefined dictionary needs to sufficiently cover all possible scenes \cite{hillar2015can}.

Our proposed algorithm, Online FISTA, offers a couple key advantages over existing compressive sensing methods in the context of SAR. First, the algorithm requires no storage of prior pulse data, as it only relies on the previously collected signal and sufficient statistics, thereby reducing the computational burden on memory-constrained systems. Second, because the image is reconstructed online, imaging parameters (such as RF-band, look direction, and pulse intervals) can be dynamically adjusted to improve data retrieval. Finally, by online reconstruction of the image during pulsing, Online FISTA can be incorporated into real-time ATR instead of waiting for data collection to complete, allowing for faster target identification.

The outline of the remainder of the paper is as follows. Section~\ref{sec:radar-imaging} introduces the radar imaging problem in the context of mono-static SAR. Section~\ref{sec:sparse-sar} introduces our proposed algorithm and its application to the sparse SAR problem. Section~\ref{sec:experiments} presents the conducted experiments and results on several sample scenes. Finally, Section~\ref{sec:conclusion} summarizes our findings and future direction of this topic.

\section{Radar Imaging}
\label{sec:radar-imaging}
To simplify the problem, we assume the start-stop approximation and the far-field assumption hold \cite{carrara1995soptlight}.

Let $\Psi:\R^2\rightarrow \R$ be the ground topography. We use $\bx\in\R^2$ and $\xx\in\R^3$ to denote the location of the target in $\R^3$, where $\xx=\begin{bmatrix}\bx,\Psi(\bx)\end{bmatrix}$. We also denote the trajectory of our antenna to be $\gamma:\begin{bmatrix}s_0,s_n\end{bmatrix}\rightarrow\R^3$, where for each slow-time instance $s_n$ we are at some location in 3D-continuous space. For our aspect of the SAR problem, we aim to recover the reflectivity function $\rho:\R^2\rightarrow\R$ of a scene of interest. To do this, we pulse over the scene and receive the back-scattered signal $d(t,s_n)$ at fast-time (range) $t$ and slow-time (pulse number) $s_n$, modeled as
\begin{equation}
    d(t,s_n)=\mathcal F[\rho](t,s_n)+\varepsilon_n(t,s_n),\label{eq:bsc}
\end{equation}
where $\mathcal F(\cdot)$ is a Fourier Integral Operator (FIO) \cite{duistermaat1996fourier, wang2014bistatic, krishnan2011synthetic, yarman2007bistatic, intes2004diffuse} representing the SAR forward model and $\varepsilon_n(t,s_n)$ represents additive noise at pulse $n$. We define the FIO for SAR under the Born approximation as follows
\begin{equation}
    \mathcal F[\rho](t,s_n):=\int e^{-j\omega\phi(t,s_n,\xx)}A(\omega,s_n,\bx)\rho(\bx)d\omega d\bx,\label{eq:dtsn}
\end{equation}
where $\omega$ is the fast-time frequency, $\phi$ is a phase function, and $A$ is an amplitude function that varies slowly with respect to $\omega$ \cite{mason2017deep}. Without loss of generality we assume $A\equiv1$ in the context of our problem and define phase as
\begin{equation}
    \phi(t,s_n,\xx)=t-\frac{2}{c}|\gamma(s_n)-\xx|,\label{eq:phase}
\end{equation}
where $c$ is the speed of light in free-space, $\gamma(s_n)\in\R^3$ is the position of the platform at pulse $n$, and $|\cdot|$ denotes the Euclidean norm. To obtain a discrete model of \eqref{eq:bsc}, we discretize the fast-time into $N_r$ samples $d(t_m,s_n)$ and organize it into a vector $\dd{n}\in\C^{N_r}$. Next we discretize $\xx$ into $N$ values, $\bx_i,\;i=1,...,N$, such that $\rho(\bx_i)$ is organized into a vector $\br\in\C^N$. Let $\mathbf F$ be the resulting matrix representing a discrete approximation to $\mathcal F$. Then we write
\begin{equation}
    \dd n = \mathbf F\br+\varepsilon_n,\quad\varepsilon_n\sim\mathcal N(0,\RR_n),
\end{equation}
where $\RR_n\succ0$ is the (per-pulse) noise covariance matrix.

\section{Sparse SAR}
\label{sec:sparse-sar}

Our objective is to reconstruct $\br$, through estimation of a set of sparse coefficients. To do so we discuss the formation of our dictionary, the sparse SAR image reconstruction problem, and our proposed online FISTA algorithm.

\subsection{Dictionary Model}
We define our over-complete dictionary $\HH\in\R^{N\times M}$ to be comprised of $M$ atoms with each atom being denoted by $N$ pixels (where $M>N$ signifies over-completeness). Each atom corresponds to a binary edgelet ($1$ if part of the edgelet, $0$ otherwise) defined in an image space of size $\R^{\sqrt{N}\times\sqrt{N}}$, with a rotation of $\psi\in[0,2\pi)$, length of $l\in\{1,2,...,\sqrt{N}-1\}$, and a centered origin of $(x,y), \;x,y\in\{0,1,...,\sqrt{N}-1\}$. Several sample (non-vectorized) atoms can be seen in Figure~\ref{fig:edgelets}.

\begin{figure}
    \centering
    \includegraphics[width=1\linewidth]{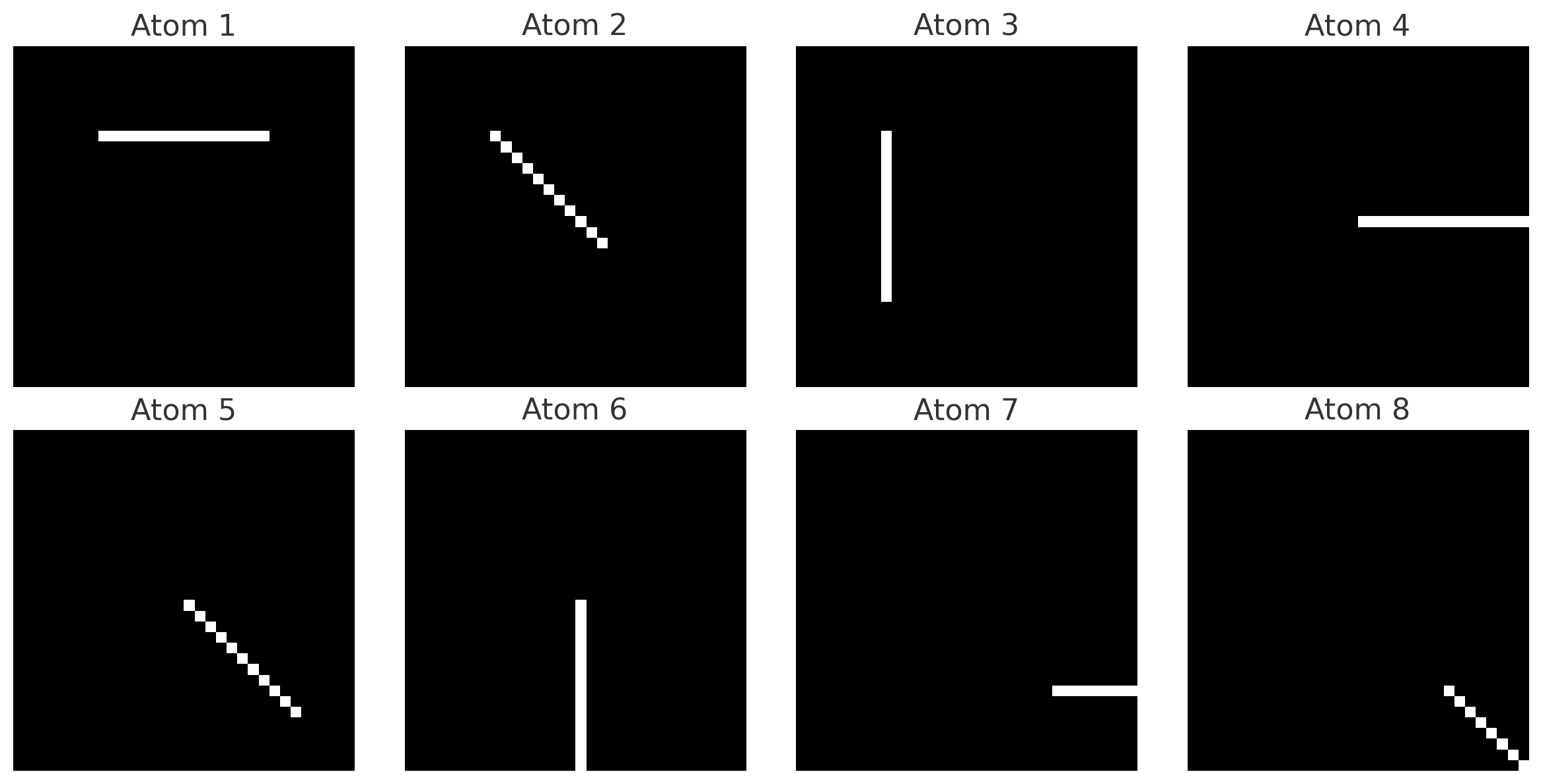}
    \caption{Example of non-vectorized edgelet based atoms.}
    \label{fig:edgelets}
\end{figure}

Given a sufficiently large dictionary $\HH\in\R^{N\times M}$ and a discretized kernel of $\mathbf{F}\in\C^{N_r\times N}$, we can reconstruct the vector of back-scattered signals at pulse $n$, $\dd n\in\C^{N_r}$, as:
\begin{equation}
    \dd n =\GG\cc+\varepsilon_n,\quad\varepsilon_n\sim\mathcal C\mathcal N(0, \RR_n),\label{eq:dn_reconst}
\end{equation}
where
\begin{equation}
    \GG=\FF\HH,\quad\GG\in\C^{N_r\times M}\label{eqn:G}
\end{equation}
is the per-pulse forward operator and $\cc\in\R^M$ is the sparse coefficient vector. Measurements across pulses are conditionally independent given $\cc$.

\subsection{SAR Image Reconstruction as a Sparse Signal Recovery Problem}
We define our objective such that given an over-complete dictionary $\HH$ and a discrete FIO approximation matrix $\FF$, can we determine a sparse coefficient vector $\cc$ such that
\begin{equation}
    \br\approx\HH\cc.
\end{equation}
By collecting a set of $n$ pulses $\{\dd k, \GG_k, \RR_k\}_{k=1}^n$, we can fit an optimal $\cc$ to the data through the Weighted LASSO objective \cite{fista, garrigues2008homotopy} such that
\begin{equation}
    \mathcal J_n(\cc):=\min_\cc\frac{1}{2}\sum_{k=1}^{n}(\dd k-\GG_k\cc)^{\dagger}\RR_k^{-1}(\dd k-\GG_k\cc)+\lambda||\cc||_1.
\end{equation}
Note $(\cdot)^\dagger$ denotes the Hermitian Transpose operator. Let $f(\cc)$ be defined as
\begin{equation}
f(\cc)=\frac{1}{2}\sum_{k=1}^n(\dd k-\GG_k\cc)^\dagger\RR_{k}^{-1}(\dd k-\GG_k\cc),
\end{equation}
then we define the gradient of $f(\cc)$, $\nabla f(\cc)$ to be
\begin{equation}
    \nabla f(\cc)=\sum_{k=1}^n\GG_k^\dagger\RR_k^{-1}(\GG_k\cc-\dd k).
\end{equation}
Furthermore, a valid Lipschitz constant for $\nabla f$ is
\begin{equation}
    L\geq ||\sum_{k=1}^n\GG_k^\dagger\RR_k^{-1}\GG_k||_2,
\end{equation}
where $||\cdot||_2$ denotes the spectral norm \cite{xiao2009dual, asif2010dynamic}.

\subsection{Online FISTA}
\label{subsec:fista}
We introduce Online FISTA, an online variant of FISTA \cite{fista} adapted to the SAR image reconstruction problem as shown in Algorithm~\ref{alg:online-fista}. This algorithm performs gradient updates actively after each pulse rather than running post-collection by exploiting the fact that all collected data $\dd n$ shares the same underlying $\br$, allowing us to carry our approximation of $\cc$, $\hc$, over multiple pulses rather than restarting after each pulse. To do this, we simplify the objective function to be based on the two running parameters $\AAA_n, \bb_n$ and their associated gradients $\Delta \AAA_n, \Delta \bb_n$, such that the number of per-pulse gradient iterations (inner steps) $m$ is limited. We pass in our prior coefficient approximation $\hc_{n-1}$ and sufficient statistics $\AAA_{n-1}, \bb_{n-1}, \Delta \AAA_n, \Delta \bb_n$ to Online FISTA, defining them as such
\begin{equation}
\Delta \mathbf{A}_{n} \;=\; \mathbf{G}_{n}^\HH \RR_{n}^{-1} \mathbf{G}_{n},\qquad
\Delta \bb_{n} \;=\; \mathbf{G}_{n}^\HH \RR_{n}^{-1} \dd{n},\label{eqn:delta-ab}
\end{equation}
where
\begin{equation}
\mathbf{A}_{n}= \mathbf{A}_{n-1}+\Delta \mathbf{A}_{n},\qquad \bb_{n}=\bb_{n-1}+\Delta \bb_{n}.\label{eqn:ab}
\end{equation}
We can now denote an incremental \cite{bertsekas2011incremental} bound for the smooth term for $\nabla f$ to be
\begin{equation}
    L_{n+1}\leq L_{n}+||\RR_n^{-1/2}\GG_n||_2^2.
\end{equation}
By using $\AAA_n$ and $\bb_n$ as sufficient statistics for newly received pulses, we remove the need to store $\dd n$,$\;\GG_n$, and $\RR_n$ for all $n$ greatly reducing memory capacity requirements for the system.

\begin{figure}[!hbt]
    \begin{algorithm}[H]
    \caption{Online FISTA per Pulse (Streaming Update from Pulse $n-1$ to $n$)}
    \label{alg:online-fista}
    \begin{algorithmic}[1]
    \State \textbf{Input:} previous estimate $\widehat{\cc}_{n-1}$, sufficient statistics $\mathbf{A}_{n-1}$, $\Delta \mathbf{A}_{n}$, $\bb_{n-1}$, $\Delta \bb_{n}$, inner steps $m$, step size $\tau=1/L_{n}$, previous momentum coefficient $t_{n-1}$
    \State \textbf{Initialize:} $\cc^{(0)}=\widehat{\cc}_{n-1}$, $\zz^{(1)}=\cc^{(0)}$
    \If{$n=1$}
        \State $t_{1}=1$ \Comment{(default to $1$ if first run)}
    \Else
        \State $t_{1}=t_{n-1}$
    \EndIf
    \For{$k=1,2,\dots,m$}
      \State $\mathbf{g}^{(k)} \gets \mathbf{A}_{n}\,\zz^{(k)} - \bb_{n}$ \Comment{(gradient)}
      \State $\uu^{(k)} \gets \zz^{(k)} - \tau\, \mathbf{g}^{(k)}$
      \State $\cc^{(k)} \gets \mathcal{S}_{t\tau}\!\big(\uu^{(k)}\big)$
      \Comment{$\mathcal S_{\alpha}(u)=\max\{|u|-\alpha, 0\}$} 
      \State $t_{k+1} \gets \dfrac{1+\sqrt{1+4t_k^2}}{2}$
      \State $\zz^{(k+1)} \gets \cc^{(k)} + \dfrac{t_k-1}{t_{k+1}}\big(\cc^{(k)}-\cc^{(k-1)}\big)$
    \EndFor
    \State \Return $\widehat{\cc}_{n} \gets \cc^{(m)}, t_n\leftarrow t_{k+1}$
    \end{algorithmic}
    \end{algorithm}
\end{figure}

\subsection{Memory Usage}
To determine the memory efficiency of our algorithm relative to prior methods like FISTA \cite{fista}, we compare the storage requirements of our running sufficient statistics \eqref{eqn:ab} against the storage of all previous pulses $\{\dd k,\GG_k, \RR_k\}^n_{k=1}$ required by FISTA up to a given slow-time $n$. As shown in Table~\ref{table:memory}, Online FISTA is more efficient when $M\lessapprox n\cdot N_r$. In the cases where we isolate small locations in the image to perform post-processing, such that the number of pixels $N$ and atoms $M$ are small, then Online FISTA vastly outperforms traditional FISTA in memory storage requirements as it does not scale with $n$. Conversely, should the user wish to sparsely encode a scene with large dimensions, then $\AAA_n$ will be infeasible to store on a computationally limited system with $n\cdot N_r\ll M$ making prior methods more favorable.
\begin{table}[h!]
    \renewcommand{\arraystretch}{1.5}
    \centering
    \caption{Memory Table.}
    \label{table:memory}
    \begin{tabular}{||c|c|c||}
        \hline
         & FISTA & Online FISTA\\
         \hline
         Variables & $\cc,\;\HH,\;\{\dd k,\GG_k,\RR_k\}_{k=1}^n$ & $\cc,\;\HH,\;\AAA_n,\;\bb_n$\\
         \hline
         Values Stored\footnotemark & 
         $\begin{aligned}
             &(M)(N+1)+\\&(2nN_r)(1+M+N_r)
         \end{aligned}$
         & 
         $\begin{aligned}
             &(M)(N+1)+\\
             &(2M)(M+1)
         \end{aligned}$\\
         \hline
    \end{tabular}
\end{table}
\footnotetext[1]{We assume storing complex values takes twice as much memory as storing a real valued number, and therefore multiply by $2$.}

\subsection{Exact Recovery Guarantee}
\label{subsec:sampling-complexity}
The exact recovery guarantee refers to the number of slow-time measurements we require to achieve an optimal $K$-sparse reconstruction, where a reconstruction is $K$-sparse if at most $K\ll M$ of the coefficients in $\cc$ are non-zero. At each slow-time instance $s_n$, we construct the per-pulse forward operator $\GG_n$ as shown in \eqref{eqn:G}. We organize these measurements into a vector $\bar{\GG}=\{\GG_1,\GG_2,...,\GG_n\}$. Through the Restricted Isometry Property (RIP) of order $K$ \cite{wei2010sparse, candes2008restricted, candes2008introduction}, if the following condition holds
\begin{equation}
    (1-\delta_K)||\cc||_2^2\leq||\bar{\GG}\cc||_2^2\leq(1+\delta_K)||\cc||_2^2,
\end{equation}
for all $K$-sparse vectors $\cc$ and $\delta_K\in(0, 1)$, then it is possible to recover the sparse signal. As $\delta_K\rightarrow 0$, the better the sparse signal reconstruction is \cite{wei2010sparse}.

Although $\bar{\GG}$ is unavailable to us prior to measurement collection, we can reconstruct the $K$-sparse signal with high probability \cite{baraniuk2007compressive} so long as our number of measurements $n$ exceeds the necessary sampling complexity of
\begin{equation}
    n\geq\mathcal O\left(K\log\left(\frac{N}{K}\right)\right).
\end{equation}
where $N$ represents the dimension of $\br\in\C^N$. 

However, for this compressive sensing condition to hold, randomness must be introduced into the measurement collection process \cite{wei2010sparse}. To achieve this, we perform a Bernoulli trial (with a probability of success $p_{\text{adjusted}}$) at each linearly spaced slow-time interval along the platform's trajectory to determine whether to transmit a pulse or move to the next position. This results in a randomly sampled set of measurements that satisfies the required condition.

\section{Experiments}
\label{sec:experiments}
In this section we discuss the experiments ran to validate the aforementioned algorithm as well as the results found.

\subsection{Experiment Setup}
We run our algorithm on four different simple simulated scenes to test its performance. These scenes, shown in Figure~\ref{fig:scenes}, contain a square (Scene 1), two squares (Scene 2), and lines of varying intensities both spaced and adjoined (Scenes 3 and 4).

Two different dictionary compositions were used, one for Scenes 1 and 2, and another for Scenes 3 and 4. For the first dictionary, we use a dictionary of edgelets comprised of length 4 and rotations of 0 and 90 degrees (horizontal and vertical). The second dictionary is comprised of edgelets of lengths 2, 4, and 6 and only a rotation of 0 degrees (horizontal). The reason for this difference is explained in Section~\ref{subsec:results}.

To collect data, all platforms are run along the same trajectory of a radius of 4km from the target location, 1km in the air. This trajectory performs spotlight sampling of the ground target, pulsing at randomly sampled locations from 1000 linearly spaced locations along a 0 to 2 degree angle. Should the ideal number of coefficients be reached before the platform reaches the end of its trajectory, the simulation auto-terminates. The ideal number of coefficients for each scene are 4, 8, 5, and 5 respectively. We also define the internal steps coefficient for Online FISTA to be $m=20$.

As a baseline, we use the Fourier-based method BP \cite{jakowatz2008beamforming} defined as
\begin{equation}
    \br\approx20\log_{10}\left(\frac{1}{n}\sum_{k=1}^n\FF^\dagger_k\dd{k}\right),
\end{equation}
where $\FF^\dagger_k$ is the Hermitian Transpose of the forward model at slow-time $s_k$ and $\dd k$ is the corresponding back-scattered signal. The reconstructed image is then converted to dB.

\begin{figure}
    \centering
    \includegraphics[width=1\linewidth]{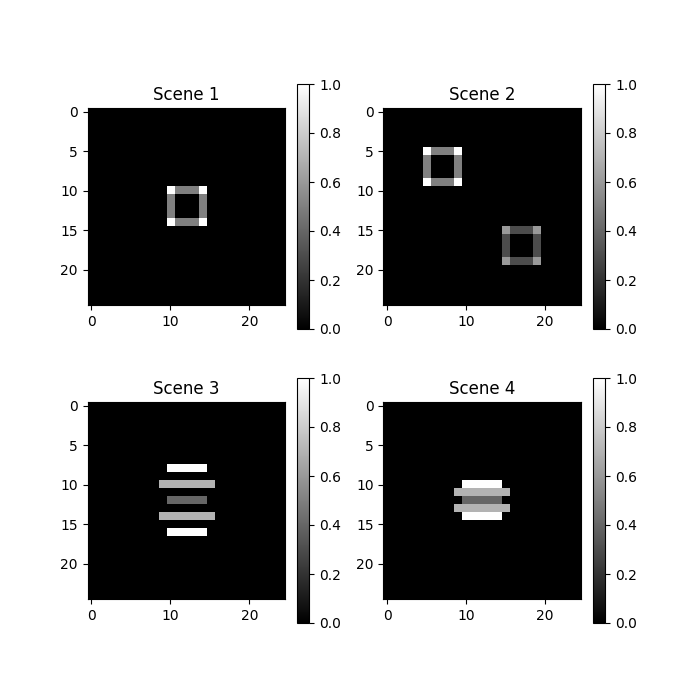}
    \caption{Experiment sample scenes.}
    \label{fig:scenes}
\end{figure}

\subsection{Results}
\label{subsec:results}

First, we review the large coefficient ($c>2\times 10^{-2},\;\forall c\in\cc$) count compared to slow-time, as show in Figure~\ref{fig:coeffs}. We can see for Scenes 1 and 2 the algorithm quickly converges to the optimal coefficient layout. For Scenes 3 and 4, it takes longer to converge as the varying lengths in edgelets causes an over-representation to the predicted scene, making it harder to recover the ideal coefficients \cite{sigg2012learning}. This shows that given a well populated dictionary, that both sufficiently represents the scene without overlapping atoms, we can have a timely and ideal sparse reconstruction \cite{hillar2015can}.

\begin{figure}
    \centering
    \includegraphics[width=1\linewidth]{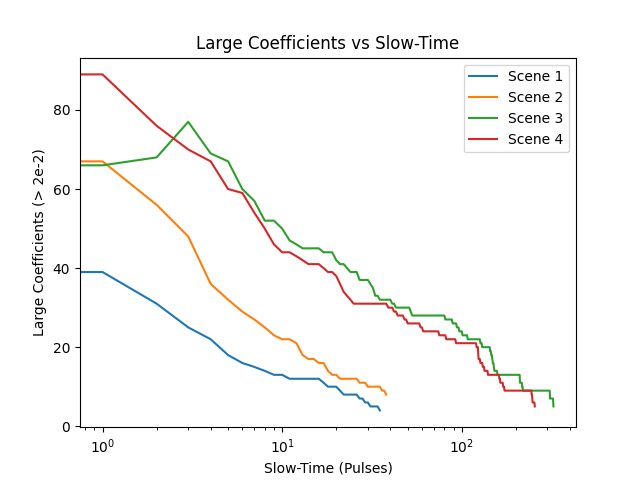}
    \caption{Large coefficients ($c>2\times 10^{-2}$) compared to slow-time over each scene.}
    \label{fig:coeffs}
\end{figure}

Second, we examine the Signal-to-Noise Ratio (SNR) \cite{gonzalez2009digital} as a function of slow-time in Figure~\ref{fig:snr} defined as
\begin{equation}
    \text{SNR}_{\text{dB}}=20\log_{10}\left(\frac{\mu_{\text{signal}}}{\mu_{\text{noise}}}\right)
\end{equation}
where $\mu_{\text{signal}}$ is the power of the signal and $\mu_{\text{noise}}$ is the power of the noise. We observe that SNR of our method averages around $70$dB once the coefficient counts converge to the optimal sparse amounts. SNR is not higher due to small remaining constants ($c\leq2\times 10^{-2}$) creating faint noise perturbations in the background, where thresholding out these small constants would greatly improve SNR due to $\mu_{\text{noise}}\rightarrow0$. We can also see the performance of Online FISTA compared to the baseline (BP) in Figure~\ref{fig:snr}. In few pulses, the SNR of Online FISTA greatly out-performs that of the baseline showcasing the performance of compressive sensing. Figure~\ref{fig:comparison} highlights this disparity, showing the reconstruction quality at just 10 pulses for each scene.

\begin{figure}


    
    \centering
    \includegraphics[width=\linewidth]{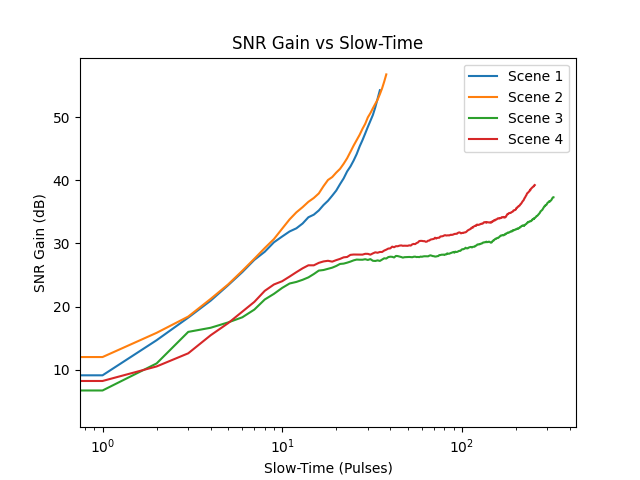}
    \caption{SNR Gain vs slow-time comparison between Online FISTA and BP. Gain is measured in the difference between the SNR of Online FISTA and BP at each slow-time instance.}
    \label{fig:snr}
\end{figure}

\begin{figure}
    \centering
    \begin{subfigure}{1\linewidth}
        \includegraphics[width=\linewidth]{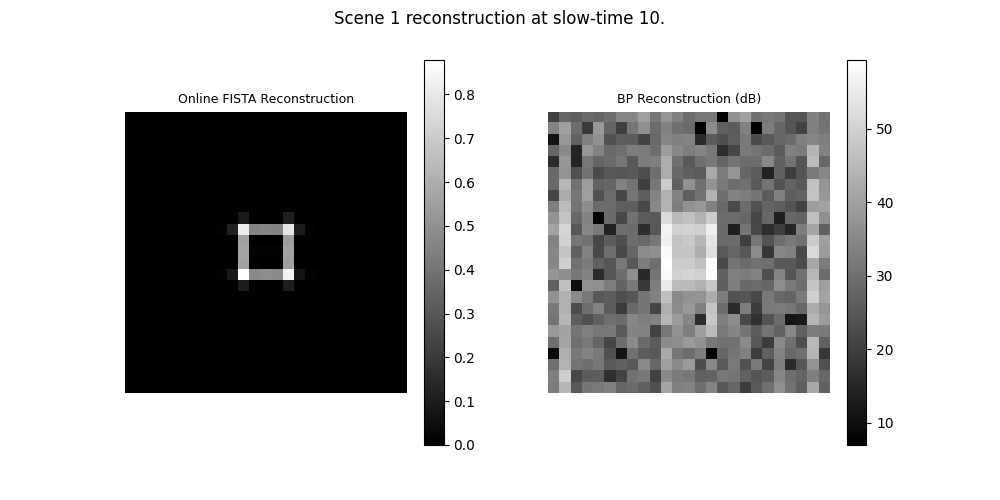}
        \caption{Scene 1 at 10 pulses.}
    \end{subfigure}

    \vspace{0.5em}
    
    \begin{subfigure}{1\linewidth}
        \includegraphics[width=\linewidth]{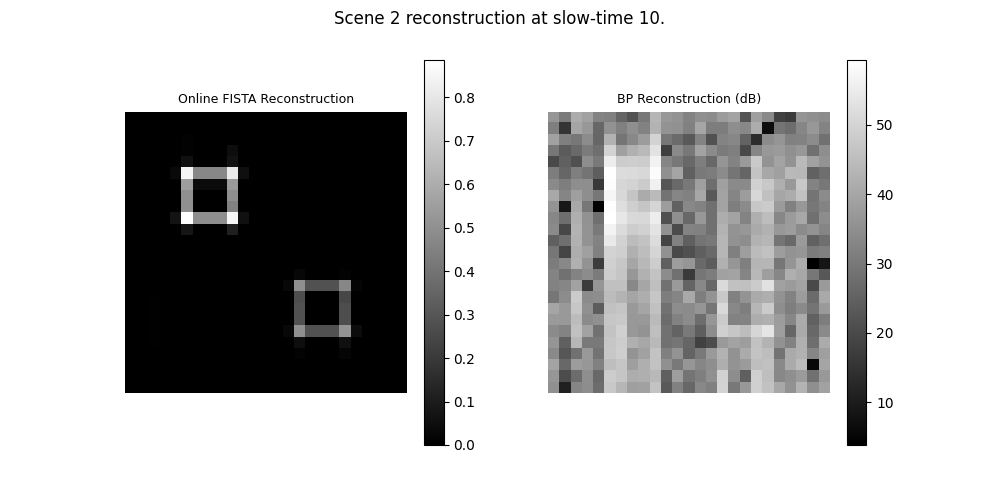}
        \caption{Scene 2 at 10 pulses.}
    \end{subfigure}

    \vspace{0.5em}
    
    \begin{subfigure}{1\linewidth}
        \includegraphics[width=\linewidth]{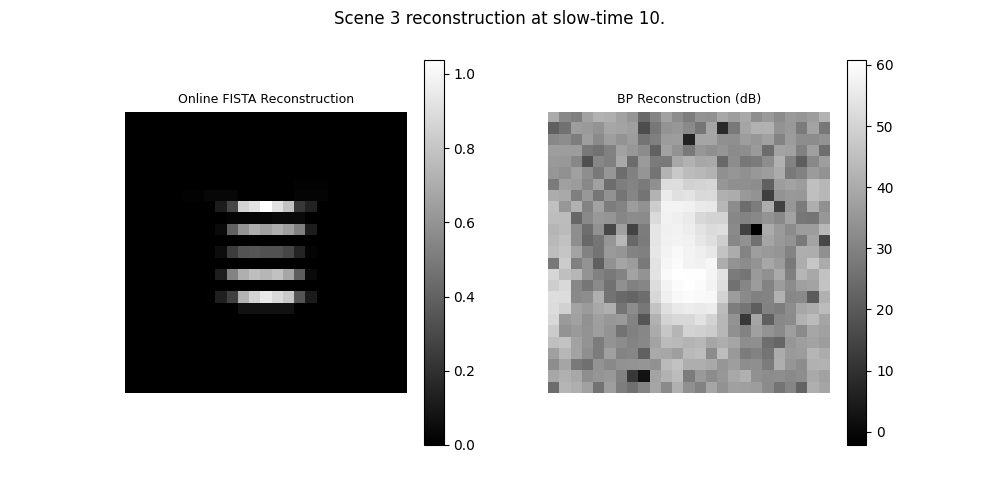}
        \caption{Scene 3 at 10 pulses.}
    \end{subfigure}

    \vspace{0.5em}
    
    \begin{subfigure}{1\linewidth}
        \includegraphics[width=\linewidth]{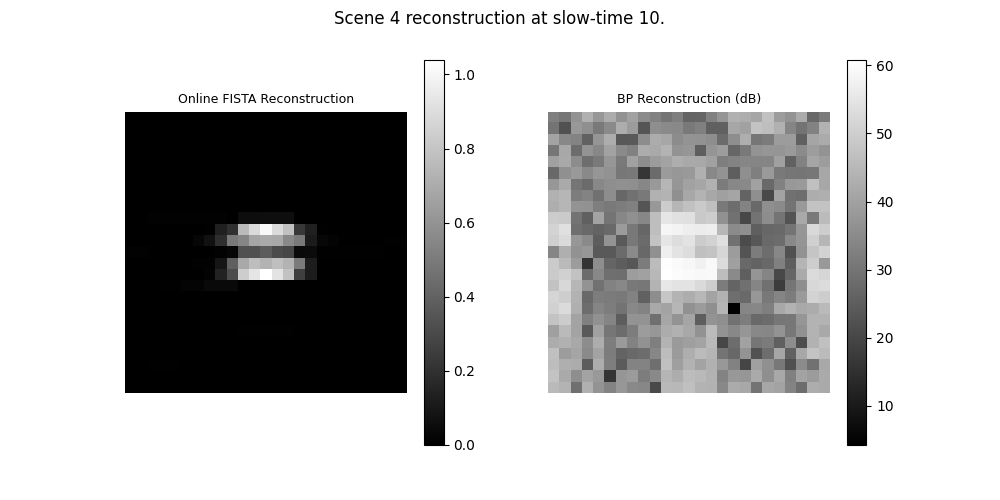}
        \caption{Scene 4 at 10 pulses.}
    \end{subfigure}
    
    \caption{Partial reconstruction comparison between Online FISTA and BP.}
    \label{fig:comparison}
\end{figure}

Thus, we have shown that both sparse representation and accurate scene reconstruction are viable through use of Online FISTA. While both require a sufficiently defined dictionary, only sparse representation is contingent on the edgelets not having significant overlap.

\section{Conclusion}
\label{sec:conclusion}

This paper has presented a new Online FISTA, an online compressive sensing algorithm designed for sparse SAR imaging. The method offers reduced memory requirements and enables online parameter tuning and online ATR by iterating FISTA at each slow-time instance rather than performing reconstruction only after full data collection. A key component of the design is the dictionary construction, which must sufficiently represent expected scene structures while minimizing overlap between edgelets to avoid suboptimal sparse encodings. We also showed that satisfactory reconstruction can be justified through the Restricted Isometry Property (RIP), which provides a probabilistic guarantee of optimal reconstruction after a sufficient number of pulses, assuming randomness is introduced into the sampling process. Experimental results demonstrate that the proposed method requires far fewer pulses to reach optimal coefficients and achieves high-quality reconstructions with only a small number of pulse samples, outperforming a Fourier-based baseline (BP).

\bibliographystyle{IEEEtran}
\bibliography{bibi}

\end{document}